\documentclass[sigconf,nonacm]{acmart}
\AtBeginDocument{%
  }

\usepackage{balance}

\setcopyright{none}
\copyrightyear{2026}
\acmYear{2026}

\begin{document}

\title{A Hybrid State-Space Approach for Census-Tract Population Estimation}

\author{Jackson R. Ye}
\email{ry135@rutgers.edu}
\affiliation{%
  \institution{Rutgers University}
  \city{New Brunswick}
  \state{NJ}
  \country{USA}
}

\author{Alexandre V. Morozov}
\email{morozov@physics.rutgers.edu}
\affiliation{%
  \institution{Rutgers University}
  \city{New Brunswick}
  \state{NJ}
  \country{USA}
}

\renewcommand{\shortauthors}{Ye and Morozov}

\begin{abstract}
Sequence models---the architecture family behind large language models and, increasingly, state-of-the-art image recognition---have redefined how machines learn from high-dimensional data. Yet population estimation from satellite imagery, a task that underpins infrastructure planning, public health, and disaster response, has scarcely benefited: leading systems still bind population to a uniform raster, disaggregating census counts onto grid cells through weighting surfaces built from ancillary data (e.g., in WorldPop and LandScan), which can introduce systematic spatial bias, and predicting population per grid cell with convolutional neural networks. In this approach, the administrative-unit structure in which the census was actually collected is discarded. We close this gap with MambaPop, which renders each administrative unit as a single polygon-masked satellite image and treats tract-level population estimation as a sequence-modeling problem over its image patches, pairing each tract image directly with its population label and eliminating the disaggregation step entirely. Built on the hybrid state-space--attention MambaVision backbone, MambaPop is, to our knowledge, the first method to learn population directly from an administrative unit's own image as well as the first to apply a state-space based (Mamba) hybrid architecture to the population estimation task. Across all $\sim$84{,}000 contiguous-US census tracts of the 2020 census, MambaPop attains a mean absolute error (MAE) of $1{,}141$ persons per tract, matching the strongest convolutional baseline (YOLOv11, MAE $1{,}122$). 
Trained only on 2010 population labels and evaluated on census data collected a decade later, MambaPop again exhibits predictive power comparable to YOLOv11 and superior to a pure-attention Vision Transformer. Finally, MambaPop is the most accurate method for predicting the population of large tracts that had been completely excluded from the training data. Taken together, our results establish sequence modeling as a competitive computational approach for tract, county and state-level population estimation, enabling low-cost population monitoring from publicly available satellite imagery alone.
\end{abstract}

  
\begin{teaserfigure}
    \centering
    \includegraphics[width=\textwidth]{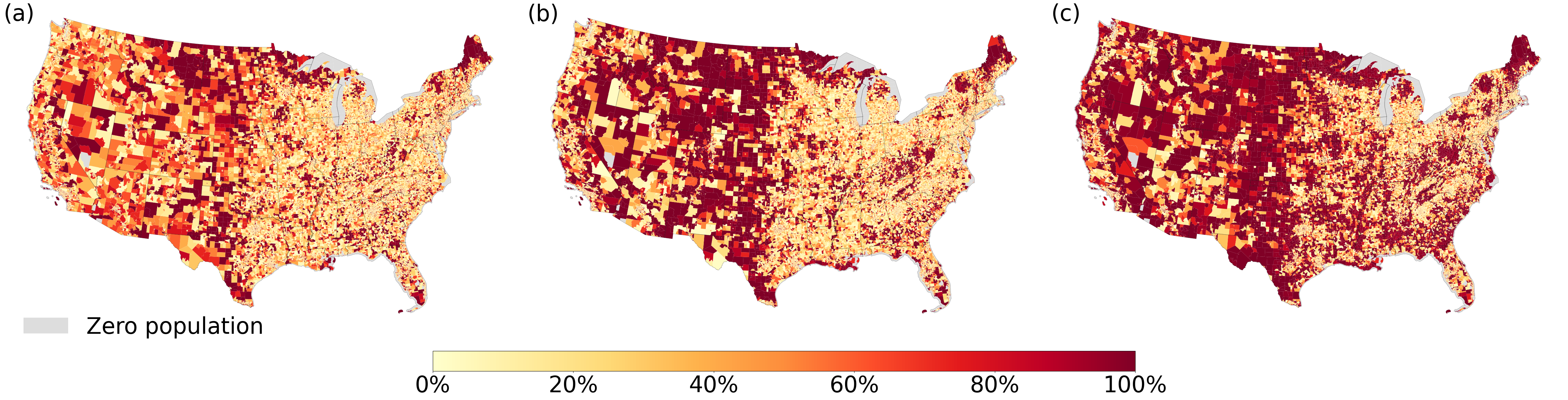}
    \caption{Predicted population percentage error, defined as $\left|\hat{y} - y\right|/y$ where $\hat{y}$ is the model prediction and $y$ is the ground-truth tract population, across the contiguous United States for (a) MambaPop, (b) YOLOv11, and (c) ViT-B/16. All models are trained on 2010 Landsat-7 imagery with SF1 census-tract labels and evaluated on the 2020 census-tract boundaries. Gray tracts have ground-truth zero population and are therefore excluded from the percentage error calculation.}
    \Description{Teaser}
    \label{fig:contiguous_us}
\end{teaserfigure}

\maketitle

\section{Introduction}
\label{sec:introduction}

Accurate, up-to-date population data underpins infrastructure planning, healthcare provisioning, disaster response, humanitarian assistance, and resource distribution~\cite{o2019importance,harper2012applications,tatem2012mapping}. Authoritative population counts come from national censuses~\cite{o2019importance,christen2023thirty}, which are expensive to conduct and are typically carried out only once per decade or less, leaving a multi-year window in which planners and responders must fall back on surveys or model-based projections~\cite{uscb2020acs}. Estimating populations directly from satellite imagery---continuously acquired and globally available---is therefore an attractive complement to in-person enumeration, and advances in deep learning have made it increasingly viable~\cite{reichstein2019deep,zhu2017deep,robinson2017deep,hafner2023mapping,duan2024high,lei2024improved}.

In parallel, machine learning has been reshaped by \emph{sequence models}. Transformers~\cite{vaswani2017attention} power today's large language models and, as Vision Transformers~\cite{dosovitskiy2020image}, increasingly rival convolutional neural networks (CNNs) on image recognition, while structured state-space models (SSMs) such as Mamba~\cite{gu2021efficiently,gu2023mamba} extend this paradigm to long sequences at linear cost. Population estimation from imagery, however, has not yet fully benefited from this shift. Operational pipelines remain dominated by grid-based approaches such as WorldPop~\cite{tatem2017worldpop} and LandScan~\cite{dobson2000landscan}, which partition imagery into a uniform raster and disaggregate census counts onto it using ancillary-data weighting surfaces---from expert-assigned coefficients to fitted random-forest layers~\cite{stevens2015disaggregating}. Disaggregation may introduce systematic spatial bias; moreover, it depends on the availability and quality of ancillary inputs~\cite{wardrop2018spatially} and discards the administrative-unit structure, such as counties and tracts, in which census data is actually collected and used~\cite{leyk2019spatial}. Even recent deep-learning approaches~\cite{robinson2017deep} inherit this per-grid-cell view, predicting populations at the gridded raster cell rather than the direct administrative unit level.

We argue that the most natural unit for population estimation is the administrative unit itself, and that its image is best modeled not as a fixed grid but as a \emph{sequence}. A census tract is an irregular, variable-size region; its imagery is naturally read as a sequence of image patches---precisely the input that sequence models are built to process. This reframing turns tract-level population estimation into a sequence-modeling problem and lets us pair each tract image directly with its true Summary File~1 (SF1) population label, eliminating disaggregation entirely. We instantiate it with MambaPop, built on MambaVision~\cite{hatamizadeh2025mambavision}---a hybrid backbone that combines convolutional stages with structured state-space and windowed-attention blocks. To our knowledge, this is the first approach to estimate population directly from an administrative unit's own image---without disaggregation of census counts onto a grid or subsequent aggregation of grid-cell predictions into meaningful administrative units---and the first to apply a hybrid state-space (Mamba) model to the task. We benchmark it against the architecture lineage that precedes it---the convolutional network of Robinson et al.~\cite{robinson2017deep} (VGG), a state-of-the-art convolutional model (YOLOv11), and a pure-attention Vision Transformer (ViT-B/16)---all trained on 2010 SF1 labels and evaluated a decade later on the 2020 census-tract boundaries to test generalization across decennial cycles, a key requirement for between-census estimation.

Our principal contributions are as follows:
\vspace{-0.15in}
\begin{itemize}
\item \textbf{A tract-level, administrative-unit-native dataset.} We construct a contiguous-US dataset at the census-tract level---a standard U.S. administrative reporting unit---pairing each tract's variable-size, polygon-masked satellite image directly with its authoritative SF1 population label and removing the disaggregation step entirely.
\item \textbf{Sequence modeling without gridding.} To our knowledge, we are the first to apply sequence modeling to population estimation directly at the administrative-unit level (from a unit's own image, without any gridding or disaggregation) and the first to use a hybrid state-space (Mamba) model for the task. We benchmark MambaPop under a single pipeline against the convolutional (VGG, YOLOv11) and attention-based (ViT) families---the first benchmark of population prediction approaches. 
\item \textbf{Generalization across vintage and scale.} MambaPop generalizes along the two axes that govern between-census utility: trained on a single decennial census, it exhibits considerable predictive power across the ensuing inter-decennial gap. In addition, MambaPop is capable of predicting population of the oversized tracts that were entirely withheld from training, demonstrating the ability of our approach to extrapolate in both time and space.
\end{itemize}

\section{Related Work}

Operational population estimation has been historically dominated by gridded approaches that disaggregate census counts onto a fine spatial grid using ancillary data such as land cover, road networks, and nighttime lights. WorldPop~\cite{tatem2017worldpop}, LandScan~\cite{dobson2000landscan}, and the CIESIN Gridded Population of the World v4~\cite{doxsey2015taking} are canonical examples, providing sub-kilometer-resolution population surfaces worldwide. More recent WorldPop releases additionally constrain disaggregation to building footprints~\cite{bondarenko2020census}. Random-forest-based dasymetric weighting~\cite{stevens2015disaggregating} is the dominant statistical approach in this paradigm. These widely-used techniques inherit limitations from the disaggregation step: the heuristics for distributing census counts within an administrative unit may introduce systematic spatial bias and depend on the availability and quality of ancillary inputs.

Previous work has applied deep learning directly to satellite imagery for population estimation. Robinson et al.~\cite{robinson2017deep} were among the first to train a convolutional neural network (a modified VGG-A) on Landsat-7 imagery, framing the task as 17-class log-scale binning followed by class-to-continuous conversion at evaluation, with fine-cell predictions aggregated back to the county level. POMELO~\cite{metzger2022fine} extended this direction by combining high-resolution satellite imagery with open geodata for fine-grained gridded population mapping. Hafner et al.~\cite{hafner2023mapping} applied a CNN to Sentinel-2 imagery for high-resolution urban-population mapping in Kigali. Duan et al.~\cite{duan2024high} used SDGSAT-1 glimmer imagery for the Greater Bay Area, and Lei et al.~\cite{lei2024improved} combined nighttime light, 3D building data, points-of-interest, and land-use within a multiscale geographically weighted regression model for China. Hu et al.~\cite{hu2019mapping} applied a similar CNN-based approach to map missing population in rural India from satellite imagery.

More recent work brings learned representations and sequence models to the task but still uses grid-based representations: Huang et al.~\cite{9419720} and Neal et al.~\cite{neal2021censusindependentpopulationestimationusing} rely on population regression over fixed grid cells, while the transformer model of Yan et al.~\cite{YAN2025101638}---the first transformer reported for population estimation---disaggregates county census onto a 500\,m grid before regression. Thus, the unit of prediction is a grid cell rather than an administrative unit, with labels still based on disaggregation or aggregation; no prior method predicts population directly at the administrative-unit level, and although state-space models have been employed in remote sensing in the context of classification and change detection tasks~\cite{10542538,10565926}, they have not yet been applied to population estimation. In this work, we directly address these gaps.

Adjacent lines of work use satellite imagery for related socioeconomic prediction tasks. Tiecke et al.~\cite{tiecke2017mapping} mapped global population at high resolution via building-footprint detection from sub-meter imagery, an approach subsequently adopted by large-scale operational systems. Jean et al.~\cite{jean2016combining} demonstrated that combining satellite imagery with machine learning can predict consumption and asset wealth in African regions where survey data is scarce, establishing the broader paradigm of satellite-imagery-based socioeconomic estimation~\cite{yeh2020using,burke2021using}. Zhu et al.~\cite{zhu2019so2sat} introduced the So2Sat LCZ42 dataset, a large-scale Sentinel-1/2 benchmark for local climate-zone classification that has informed dataset-design practices for satellite-imagery deep learning.

\section{Dataset}
\label{sec:dataset}

We introduce a new GIS population dataset covering the contiguous United States that integrates publicly available satellite imagery with administrative-level population data at the census tract level.

\subsection{Landsat 7 Surface Reflectance Data}
Landsat 7~\cite{usgs2023landsat}, launched by NASA and the U.S. Geological Survey (USGS) in 1999, is part of the Landsat program which provides long-term Earth observation data critical for environmental monitoring and land use analysis. The Surface Reflectance Tier 1 dataset used in this work consists of atmospherically corrected reflectance values derived from the Enhanced Thematic Mapper Plus (ETM+) sensor~\cite{gee2025landsat7}. These reflectance values represent the fraction of incoming solar radiation that is reflected from the Earth's surface and are corrected for effects such as atmospheric scattering and absorption.

The data includes 20 bands~\cite{gee2025landsat7sr}, including red, green, and blue (RGB), infrared, atmospheric opacity, and cloud quality, with six primary spectral bands (Bands 1--5 and 7) and additional bands for metadata, masks, and processing byproducts. All bands are provided at a 30-meter spatial resolution, making the dataset suitable for regional- to national-scale analysis while balancing spatial detail and computational feasibility. For this study, we extract only the RGB bands (Bands 3, 2, 1) to simulate the minimal input condition and reduce model complexity.

All imagery used in this study is drawn from Landsat~7 acquisitions within calendar year 2010, matching the 2010 census labels used for training; the cross-decade evaluation (Section~\ref{sec:results}) uses imagery acquired within calendar year 2020, paired with the 2020 census-tract boundaries. The imagery is preprocessed to remove clouds, shadows, and water using the quality assurance (QA) band. We apply percentile-based min-max normalization (2nd to 98th percentile) to mitigate extreme values and standardize the pixel range across all images. Each census tract is associated with a set of RGB image patches extracted from the Landsat 7 tiles that geographically overlap with the tract boundary. These patches form the input sequences for the model.

\subsection{Census Tract Population Labels (P001001 / SF1)}
We use population data from the P001001 field in the 2010 U.S. Census Summary File 1~\cite{uscb2010sf1} (SF1), which reports total population at the census tract level. This file provides a direct, authoritative label for each geographic unit in our dataset.

Census tracts are selected across the contiguous United States, and each tract's total population count is matched to the corresponding satellite imagery patches covering that tract. The assumption is that the imagery visually encodes enough demographic-related information (e.g., building density, land use) to infer the tract-level population.

\subsection{Tract-Level Image Construction}
To construct tract-level image inputs, we begin by aligning Landsat 7 RGB imagery with U.S. census tract boundaries defined by the TIGER/Line shapefiles~\cite{uscb2010tiger}. These shapefiles are obtained from the U.S. Census Bureau and provide high-resolution polygon geometries for all census tracts in the contiguous United States.

We use GDAL~\cite{GDAL/OGR} tools, specifically the \texttt{gdal\_translate} commands, to process and extract imagery corresponding to each tract boundary. The process involves the following steps:
\begin{enumerate}
    \item Landsat tiles are first reprojected to match the coordinate reference system of the TIGER/Line shapefiles. We then crop each image to the bounding box of its corresponding tract.
    \item A mask is applied using the tract polygon such that pixels outside the administrative boundary are set to transparent ($\alpha = 0$). This allows the model to focus exclusively on features within the defined tract area.
    \item GDAL's contrast stretching and scaling options are applied to improve the visibility of structural and land-use patterns. These operations adjust the reflectance values into a higher-contrast dynamic range, making features such as buildings, roads, and vegetation more distinguishable.
    \item The final output is a PNG image with a transparent background outside the tract boundary. This image preserves fine-grained visual details, while isolating the spatial extent of each administrative unit.
\end{enumerate}

Once extracted, each tract image is resized to a common $512\times512$ resolution and tokenized into a fixed-length sequence of patch embeddings for the MambaVision model, paired with its corresponding SF1 population label. Due to computational constraints, we limit the training data to tracts whose native images are smaller than or equal to $512\times512$ pixels.

Each tract has a single SF1 population count as ground truth. Following the classification framing introduced by Robinson et al.~\cite{robinson2017deep} and adopted in subsequent satellite-imagery population work~\cite{doda2022so2sat}, we assign discrete population-class labels to enable a classification-style training signal that handles the heavy-tailed distribution~\cite{yang2021delving} of tract populations more reliably than direct regression. We group tracts into $K=18$ population classes by one-dimensional $k$-means clustering of the SF1 counts, with class boundaries placed at the midpoints between consecutive cluster centers. Because the cluster centers track the empirical density, this procedure places narrow bins across the densely populated range and wider bins in the sparse upper tail.
The 18 class boundaries and per-class tract counts (2010 data) are listed in Appendix~\ref{app:binning} (Table~\ref{tab:binning}).
The original SF1 population range is retained alongside each bin label, so that classification predictions can be converted back to per-tract population estimates at inference (Section~\ref{sec:methodology}) and evaluated against the original counts.

\begin{figure*}[!htbp]
    \centering
    \includegraphics[width=0.85\textwidth]{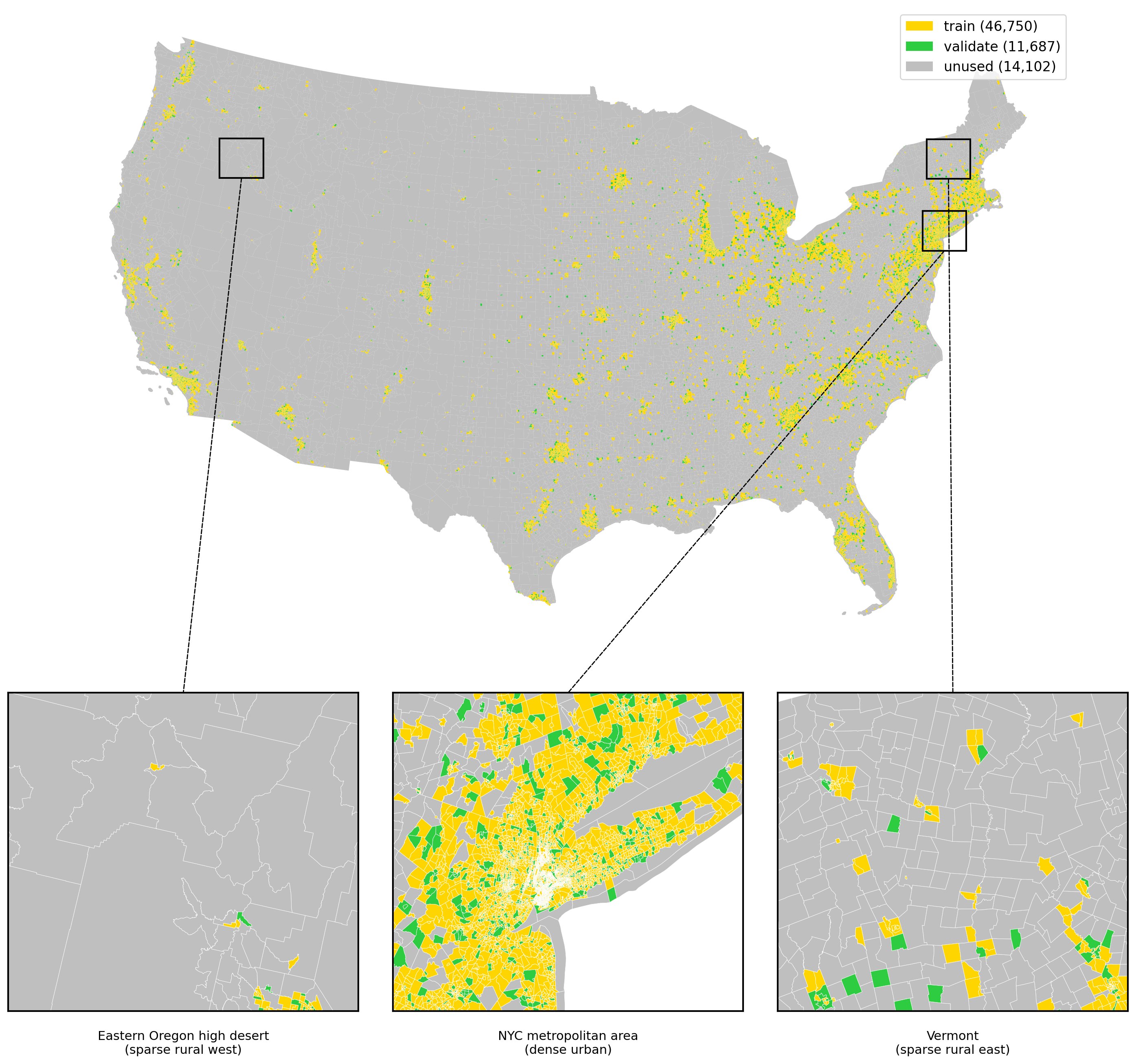}
    \caption{Geographic distribution of the 2010 train/validation split. Main map: each contiguous-US census tract colored by assignment --- train (yellow), validation (green), or unused (gray); counts shown in the upper-right legend. Insets show three representative regions illustrating the spread of train/validation coverage across rural and urban tract geometries: Eastern Oregon high desert, the New York City metropolitan area, and Vermont.
    }
    \Description{2010 train val split}
    \label{fig:train_val_split_2010}
\end{figure*}

This comprehensive image-tract construction pipeline ensures that each training example is a high-quality, visually meaningful, semantically aligned input-label pair, ready for both regression and classification modeling tasks.
Figure~\ref{fig:train_val_split_2010} shows the geographic distribution of the resulting train/validation split across the contiguous United States, with insets highlighting how the split spans both sparse rural and dense urban tract geometries.

\section{Methodology}
\label{sec:methodology}

The goal of our research is to develop a scalable and accurate framework for population estimation using only tract-level RGB imagery derived from the Landsat 7 dataset. Our methodology leverages the dataset described in Section~\ref{sec:dataset} to train an SSM image model to predict population counts directly from the satellite imagery, without requiring pixel-level or gridded population labels.

\subsection{Mamba block}
Structured state space models (SSMs), originally introduced for continuous-time sequence modeling~\cite{gu2021efficiently,gu2023mamba}, form the core sequential processing primitive in the MambaVision backbone.
The Mamba SSM block models sequential data using a continuous-time linear dynamical system~\cite{gu2020hippo,gu2021combining}, where the hidden state $\mathbf{h}(t) \in \mathbb{R}^M$ evolves based on the input $\mathbf{x}(t) \in \mathbb{R}$
and the output $\mathbf{y}(t) \in \mathbb{R}$ is computed using learnable parameters $\mathbf{A} \in \mathbb{R}^{M \times M}$, $\mathbf{B} \in \mathbb{R}^{M \times 1}$, and $\mathbf{C} \in \mathbb{R}^{1 \times M}$ as:

\begin{equation}
\mathbf{h}'(t) = \mathbf{A}\mathbf{h}(t) + \mathbf{B}\mathbf{x}(t)
\end{equation}
where $\mathbf{h}'(t)$ denotes the time derivative of the hidden state, and

\begin{equation}
\mathbf{y}(t) = \mathbf{C}\mathbf{h}(t)
\end{equation}

In practice, this continuous-time formulation is implemented in discretized form via a zero-order hold transformation~\cite{gu2023mamba}, yielding the discrete-time recurrence $\mathbf{h}_t = \bar{\mathbf{A}}\mathbf{h}_{t-1} + \bar{\mathbf{B}}\mathbf{x}_t$ and $\mathbf{y}_t = \mathbf{C}\mathbf{h}_t$ that runs inside each MambaVision Mixer block. Recent work establishes duality between structured state-space recurrences and attention, providing a theoretical link between the two computational primitives that MambaVision combines~\cite{dao2024transformers}.

\subsection{Self-attention}
Recent hybrid architectures combine structured state-space recurrence with windowed self-attention to leverage the complementary strengths of both primitives~\cite{hatamizadeh2025mambavision}.

Self-attention~\cite{vaswani2017attention}, the standard mixing primitive (i.e., information aggregation across token positions) in transformer architectures, computes contextualized token representations by weighting pairwise interactions across an input sequence:

\begin{equation}
\text{Attention}(\mathbf{Q}, \mathbf{K}, \mathbf{V}) = \text{softmax}\left(\frac{\mathbf{Q}\mathbf{K}^T}{\sqrt{d_k}}\right)\mathbf{V}
\end{equation}

where $\mathbf{Q = XW^Q}$, $\mathbf{K = XW^K}$, and $\mathbf{V = XW^V}$ are the query, key, and value matrices respectively, $\mathbf{X} \in \mathbb{R}^{n \times d}$ denotes the sequence of $n$ input token embeddings, and $d_k$ is the dimension of the key vectors. While both SSM recurrence and self-attention have individually demonstrated strong performance in vision tasks, their combination within a single backbone for administrative-unit-level prediction has so far remained unexplored.

\subsection{Model Architecture}
We employ MambaVision~\cite{hatamizadeh2025mambavision}---part of a recent wave of state-space-based vision backbones~\cite{zhu2024vision,liu2024vmamba}---as the core of our population estimation model. MambaVision is a hybrid vision backbone whose deeper stages are built around the MambaVision Mixer block, a structured state-space module derived from Mamba that processes token sequences with linear time and memory complexity. Unlike pure transformers, which rely on quadratic-cost global self-attention, and pure CNNs, which lack long-range mixing, the MambaVision Mixer combines input-dependent state-space recurrence with bounded-window context, enabling efficient modeling of long-range dependencies in the moderate-sized training regimes where standard Vision Transformers (ViTs)~\cite{dosovitskiy2020image} are sample-inefficient and prone to overfitting. Mamba's lightweight architecture and low memory footprint further support scaling population estimation to national-level datasets using high-resolution imagery.

The choice of MambaVision rests on a combination that neither baseline offers by itself: its convolutional early stages supply the locality bias suited to satellite imagery---which the pure-attention ViT lacks, which may result in overfitting (Section~\ref{sec:results})---while the MambaVision Mixer mixes the spatially dispersed cues that signal population across a tract at the linear rather than the quadratic cost of global self-attention. This pairing of locality bias with efficient long-range mixing, absent from CNN-only and ViT-only alternatives, motivates our choice of the backbone. The output from the MambaVision backbone is fed into a classification head over the $K$-class bin labels defined in Section~\ref{sec:dataset} ($K=18$). At the inference step, the predicted class is converted to a continuous per-tract population estimate by taking the mean population of the tracts assigned to that bin, following the classify-then-convert pattern introduced by Robinson et al.~\cite{robinson2017deep} for deep-learning population estimation from satellite imagery. This conversion lets us train against a discrete, well-behaved loss while reporting continuous regression-style metrics in Section~\ref{sec:results}.

\subsection{Experimental setup}
\subsubsection{Dataset Statistics}
Our final dataset consists of 72,539 census tracts across the contiguous United States, covering diverse geographic and demographic conditions. The population distribution ranges from 0 to 37,452 inhabitants per tract, with a median population of 3,995. For computational efficiency, we only include tract images smaller than or equal to $512 \times 512$ pixels, resulting in a total of 58,437 tracts used in the final dataset. We partition the dataset into training (80\%), and validation (20\%) sets using stratified sampling to ensure balanced representation across population density ranges and geographic regions.

\subsubsection{Model Configuration}
\begin{figure*}[!htbp]
    \centering
    \includegraphics[width=0.95\textwidth]{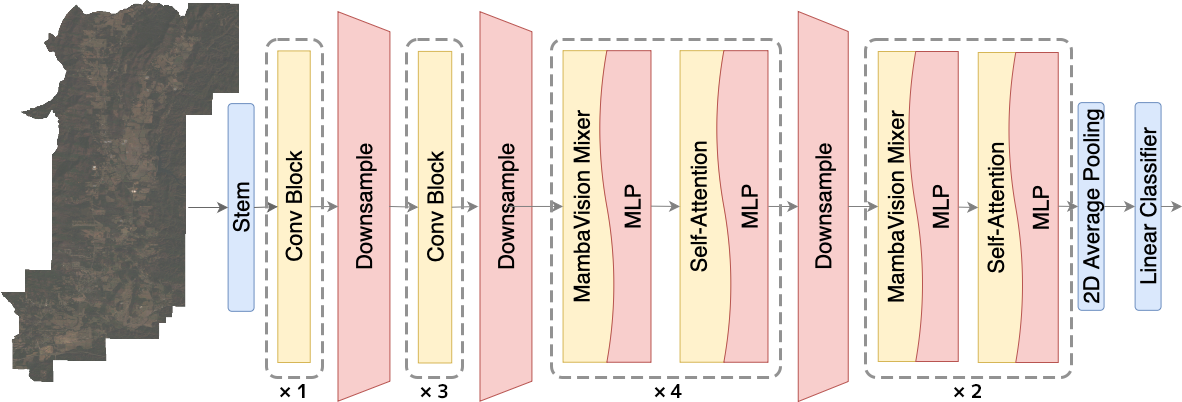}
    \caption{MambaPop architecture. The input tract image passes through a convolutional stem, convolutional residual blocks (Stages~1--2), and hybrid MambaVision Mixer and windowed self-attention blocks (Stages~3--4), followed by 2D average pooling, a linear layer, and a classification head. Blocks within each stage are applied sequentially; multipliers denote the number of repetitions of each block type.}
    \Description{Model Architecture}
    \label{fig:model}
\end{figure*}

We adopt MambaVision-T (Tiny) as our base architecture---a 4-stage hierarchical structure with layer depths $[1, 3, 8, 4]$, beginning with a convolutional stem followed by convolutional residual blocks in Stages~1 and~2. Stages~3 and~4 employ a hybrid design that interleaves MambaVision Mixer and windowed self-attention~\cite{liu2021swin} blocks: in each hybrid stage, the first half of the blocks use the MambaVision Mixer for modeling long-range dependencies, while the second half uses windowed self-attention to capture fine-grained contextual dependencies (Stage~3: 4 Mixer $+$ 4 Attention; Stage~4: 2 Mixer $+$ 2 Attention). The detailed architecture is illustrated in Figure~\ref{fig:model}.

\subsection{Training Details}
Large-scale pretraining is particularly important for sequence models. Lacking the locality and translation-equivariance biases that convolutional networks build in, pure-attention and state-space backbones are comparatively data-hungry and cannot acquire general visual representations from scratch on a dataset of our scale; ImageNet pretraining supplies these priors, which fine-tuning then adapts to population-relevant structure. We therefore initialize MambaPop, YOLOv11, and ViT-B/16 from publicly released ImageNet-1k~\cite{deng2009imagenet} pretrained checkpoints: MambaVision-T from the official MambaVision release~\cite{hatamizadeh2025mambavision}, YOLOv11n (the nano classification variant) from the Ultralytics release, and ViT-B/16 from \texttt{torchvision} (\texttt{IMAGENET1K\_V1} weights), with positional embeddings interpolated from the native 224$\times$224 grid to our 512$\times$512 input following the standard \texttt{torchvision} procedure. VGG11 is trained from random initialization, as no pretrained checkpoint suitable for the Robinson et al.~\cite{robinson2017deep} modified VGG-A architecture is publicly available. For all models, the classification head is reinitialized for the $K$-class population-bin output, and all backbone parameters are unfrozen and updated jointly with the head.

MambaPop is trained end-to-end using the prepared dataset of (satellite imagery, population class) pairs. The model parameters are optimized using AdamW~\cite{loshchilov2017decoupled} with a cosine learning-rate schedule~\cite{loshchilov2017sgdr}. The overall dataset is partitioned into training and validation sets: the training set is used to update model weights, and the validation set is reserved for final unbiased evaluation. All models are trained on 2010 SF1 labels; the 2020 census-tract boundaries are held out entirely and used for the temporal-generalization experiments in Section~\ref{sec:results}. 

\vspace{-8pt}
\subsection{Baseline Comparisons}
We benchmark MambaPop against three baselines that span the architecture families preceding it, each run under the same pipeline as MambaPop (Section~\ref{sec:methodology})---identical tract images, the same population-bin labels, and the same classify-then-convert inference---so that performance differences reflect the backbone rather than the training setup. VGG11~\cite{robinson2017deep, simonyan2014very}, the convolutional network of Robinson et al., is a classical fully convolutional classifier with global average pooling and a prediction head, long used for spatial feature extraction in gridded population estimation. YOLOv11 is a modern, widely used convolutional model representing the state-of-the-art CNN family. ViT-B/16~\cite{dosovitskiy2020image} is a pure-attention Vision Transformer that tokenizes the image into non-overlapping $16 \times 16$ patches and mixes them with global self-attention; apart from this patch-embedding layer, it has no hierarchical convolutional stem or stages, and thus lacks the locality bias that benefits the CNN and hybrid backbones in our moderate-data regime. It serves as the attention-only sequence-model baseline against MambaVision's hybrid state-space-attention design.

\subsection{Evaluation Metrics}
We evaluate model performance using several metrics to comprehensively assess both average prediction accuracy and the model's ability to explain variance in population estimates.

We report the Mean Absolute Error (MAE) and Median Absolute Error (MedAE) to summarize typical absolute deviations between predicted and actual values:
\begin{eqnarray*}
\mathrm{MAE} &=& \frac{1}{n} \sum_{i=1}^{n} \left| y_i - \hat{y}_i \right|, \\
\mathrm{MedAE} &=& \mathrm{median} \left( \left| y_i - \hat{y}_i \right| \right).
\end{eqnarray*}
Here, $y_i$ and $\hat{y}_i$ denote the observed and predicted populations of tract $i$, respectively, and $n$ is the number of tracts in the evaluation set; the same notation is used in all subsequent metrics.

We also report the coefficient of determination (\(R^2\)),  where \(\mathrm{RSS}\) is the residual sum of squares, \(\mathrm{TSS}\) is the total sum of squares, and \(\bar{y}\) is the mean of the observed values:
\[
R^2 = 1 - \frac{\mathrm{RSS}}{\mathrm{TSS}}
= 1 - \frac{\sum_{i=1}^n (\hat{y}_i - y_i)^2}{\sum_{i=1}^n (y_i - \bar{y})^2}.
\]
\(R^2\) quantifies the proportion of variance in the target variable explained by the model, compared to the variance of the actual values around their mean.

We also use the Root Mean Squared Error (RMSE), defined as:
\[
\mathrm{RMSE} = \sqrt{\frac{1}{n} \sum_{i=1}^{n} \left( y_i - \hat{y}_i \right)^2}~.
\]

Finally, we report the Mean Absolute Percentage Error (MAPE), defined as the mean absolute relative error expressed as a percentage:
\[
\mathrm{MAPE} = \frac{100}{n} \sum_{i=1}^{n} \left| \frac{y_i - \hat{y}_i}{y_i} \right|.
\]
Because the per-tract denominator $y_i$ is small for sparsely populated tracts, MAPE is highly sensitive to such tracts and can exceed 100\%; we therefore list it alongside the absolute-error metrics rather than in isolation.

\section{Results}
\label{sec:results}

\subsection{Population Estimation Performance}

We evaluate MambaPop and the baseline algorithms in two settings: an in-year setting on the 2010 held-out validation split and a temporal-generalization setting on the 2020 census-tract boundaries, with all models trained on the 2010 SF1 train split (see Section~\ref{sec:methodology}). Table~\ref{tab:model_m} presents both settings.

\begin{table*}[htb!]
\centering
\begin{tabular}{lccccc}
\toprule
\textbf{Model} & \textbf{Mean AE} & \textbf{Median AE} & \textbf{R\textsuperscript{2}} & \textbf{RMSE} & \textbf{MAPE(\%)} \\
\midrule
\multicolumn{6}{l}{\textit{2010 (held-out validation split, in-year)}} \\
MambaPop                            & 1177  & 931   & 0.35    & 1595.3  & 174.2 \\
YOLOv11                             & 1127  & 869   & 0.39    & 1548.8  & 126.2 \\
ViT-B/16                            & 1108  & 827   & 0.38    & 1560.6  & 114.7 \\
VGG11                               & --    & --    & --      & --      & --    \\
\midrule
\multicolumn{6}{l}{\textit{2020 (held-out evaluation year, temporal generalization)}} \\
MambaPop                            & 1141  & 906   & 0.21    & 1522.8  & 162.5 \\
YOLOv11                             & 1122  & 863   & 0.21    & 1516.7  & 145.2 \\
ViT-B/16                            & 1263  & 926   & $-0.05$ & 1754.0  & 124.9 \\
\bottomrule
\end{tabular}
\caption{Performance on the 2010 held-out validation split (in-year) and the 2020 census-tract evaluation set (temporal generalization). All models are trained on the 2010 SF1 train split; for each setting we report Mean AE, Median AE, $R^2$, RMSE, and MAPE. VGG11 failed to converge; consequently, dashes (--) mark the metrics that could not be reported.}
\label{tab:model_m}
\end{table*}

\begin{figure*}[htbp]
    \centering
    \includegraphics[width=1.0\textwidth]{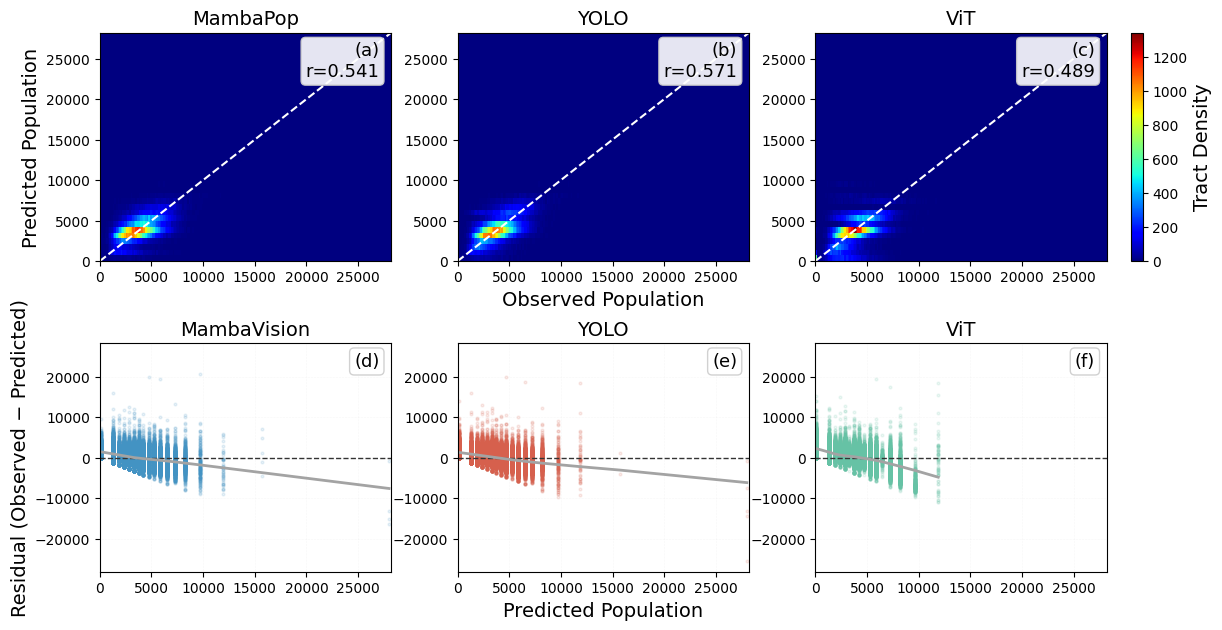}
    \caption{Predicted vs.\ Observed Population and residual diagnostics for MambaPop, YOLOv11, and ViT-B/16. Top row (a--c): density heatmaps of Predicted vs.\ Observed; the dashed line is the identity ($y = x$) and the Pearson correlation $r$ is displayed in each panel. Bottom row (d--f): residuals (observed $-$ predicted) vs.\ predicted values; the dashed line marks zero residual and the gray curve is the LOWESS (locally weighted scatterplot smoothing) trend.}
    \Description{scatter_residual and predict vs observed}
    \label{fig:scatter_residual}
\end{figure*}

On the 2010 in-year validation split, the three modern backbones perform comparably: MambaPop (MAE 1{,}177, R\textsuperscript{2}~$=0.35$), YOLOv11~\cite{khanam2024yolov11overviewkeyarchitectural} (MAE 1{,}127, R\textsuperscript{2}~$=0.39$), and ViT-B/16 (MAE 1{,}108, R\textsuperscript{2}~$=0.38$) all capture meaningful population structure from satellite imagery, while VGG11~\cite{simonyan2014very} fails to converge. The decisive differences emerge under temporal generalization. Carried forward to the held-out 2020 boundaries, MambaPop (MAE 1{,}141, R\textsuperscript{2}~$=0.21$) and YOLOv11 (MAE 1{,}122, R\textsuperscript{2}~$=0.21$) both retain positive predictive power, whereas ViT-B/16 collapses from R\textsuperscript{2}~$=0.38$ in-year to $-0.05$ on the 2020 boundaries---no better than predicting the population mean. Thus the convolutional (YOLOv11) and hybrid state-space (MambaPop) backbones generalize across the decennial gap, while the pure-attention ViT's absolute estimates do not. Figure~\ref{fig:scatter_residual} complements these quantitative results with density heatmaps of predicted vs.\ actual populations (panels a--c) and residual diagnostics (panels d--f) for MambaPop, YOLOv11, and ViT-B/16 on the 2020 set.

In the density heatmaps, MambaPop (panel~a) and YOLOv11 (panel~b) both concentrate their predictions along the identity diagonal across the low-to-mid population range, consistent with their comparable Pearson correlations ($r = 0.541$ and $r = 0.571$) and positive $R^2$ on the 2020 set. ViT-B/16 (panel~c) instead clusters its predictions in the low-population range and fails to predict high-population tracts, under-predicting them severely; despite a Pearson correlation ($r = 0.489$) close to the other two, this failure to span the full population range results in the negative 2020 R\textsuperscript{2} value (Table~\ref{tab:model_m}). The same shortfall is visible in the nationwide map of population prediction errors (Figure~\ref{fig:contiguous_us}), where ViT's largest errors fall on the high-population tracts that it cannot treat accurately.

The residual panels (d--f) reinforce this picture. For all three models the residuals are centered near zero through the low-to-mid population range that contains most tracts, with LOWESS trends drifting mildly negative only at the highest predicted values, where very high-population tracts are scarce. MambaPop (panel~d) spreads its predictions across the widest range of fitted values, whereas YOLOv11 (panel~e) and ViT-B/16 (panel~f) compress theirs into a narrower band. This effect is most pronounced for ViT, consistent with its inability to reach high-population tracts.

These results are best read through the architectural lens of the benchmark. On in-distribution tracts the convolutional YOLOv11 and the hybrid MambaVision backbone are comparable: both pair a strong locality bias with an efficient mechanism for aggregating the spatially dispersed features that signal population levels across a tract. The models separate only under distribution shift---across census vintages and across tract size---as discussed in detail below (cf. Figure~\ref{fig:large_area_error}).

VGG11 fails to converge on the 2010 training data. ViT-B/16 fits the 2010 training year well but fails to generalize temporally to 2020 (R\textsuperscript{2}~$=0.38 \rightarrow -0.05$): lacking the convolutional stem's locality bias, the pure-attention backbone overfits on the 2010 dataset rather than learning transferable visual structure, consistent with the sample inefficiency of attention architectures in moderate-data regimes~\cite{dosovitskiy2020image,touvron2021training}. MambaPop and YOLOv11 both retain positive R\textsuperscript{2} on the held-out 2020 boundaries (Table~\ref{tab:model_m}). However, MambaPop's hybrid design---convolutional stages for locality bias plus MambaVision Mixer and windowed self-attention for linear-cost long-range mixing---delivers superior performance in predicting populations of large, out-of-distribution tracts (Figure~\ref{fig:large_area_error}).

\subsection{Generalization to Large, Out-of-Distribution Tracts}
\begin{figure}[t]
    \centering
    \includegraphics[width=\columnwidth]{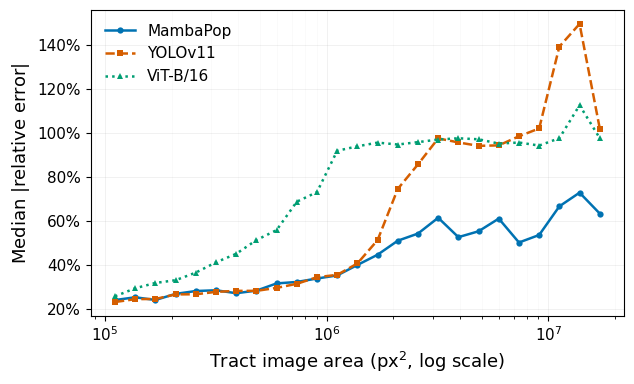}
    \caption{Median absolute magnitude of the relative error versus tract image area (log scale) on the 2020 set, for MambaPop, YOLOv11, and ViT-B/16; each marker is the median over tracts in a log-spaced area bin. The $512\times512$ training crop ($\approx$2.6$\times$10\textsuperscript{5}\,px\textsuperscript{2}) is the largest size seen in training; larger tracts are out of distribution.}
    \Description{Median relative error versus tract image area for the three backbones.}
    \label{fig:large_area_error}
\end{figure}

Training was restricted to tracts whose native images fit within $512\times512$\,pixels (px) ($\approx$2.6$\times$10\textsuperscript{5}\,px\textsuperscript{2}), so that larger tracts (predominantly rural and exurban units) are out-of-distribution at test time, differing from the training set both in content (sparse, low-density development) and in the heavier downsampling applied when their imagery is resized to the common $512\times512$ input. Notably, we evaluated all three backbones at the same $512 \times 512$ resolution, excluding any possibility of a resolution confound. Figure~\ref{fig:large_area_error} plots the median absolute magnitude of the relative error against tract image area for the three algorithms that converged on the 2020 set. Up to about $10^{6}$\,px\textsuperscript{2}, the MambaPop and YOLOv11 models are indistinguishable, with median errors near 25--40\%; in contrast, ViT-B/16 degrades immediately, producing inferior results. Beyond $10^{6}$\,px\textsuperscript{2}, MambaPop and YOLOv11 start to separate: MambaPop's error rises less rapidly, staying near 50--70\% out to the largest tracts---an order of magnitude in area beyond its training range---whereas YOLOv11's climbs past 100\% and ViT-B/16's saturates near 95--100\%. MambaPop is thus the only backbone that keeps tracking population on large tracts it never saw in training, degrading only gradually under this combined scale-and-distribution shift.

\subsection{Training Efficiency Analysis}

Table~\ref{tab:model_p} compares training cost and hyperparameters across models. MambaPop converges in 1.6~days, on par with the strongest convolutional baseline, YOLOv11 (0.95~days),
and 5.75$\times$ faster than the pure-transformer ViT-B/16, which required 9.2~days. All models were trained on a single NVIDIA GeForce RTX 3090 GPU (24\,GB).
The MambaVision Mixer's structured state-space recurrence operates in linear time with respect to the sequence length~\cite{gu2023mamba}, and the windowed self-attention in the deeper stages has per-layer cost bounded by the window size; together these keep the backbone tractable on the long token sequences from high-resolution $512\times512$ tract images, in contrast to the quadratic global attention that makes ViT-B/16 costly to train.

\begin{table*}[htbp]
\centering
\begin{tabular}{lcccccc}
\toprule
\textbf{Model} & \textbf{training time (days)} & \textbf{epochs}& \textbf{BS} & \textbf{LR} & \textbf{WD} & \textbf{DR} \\
\midrule
MambaPop                            & 1.6   & 320       & 32    & 5e-5  & 1e-2  & 0.2   \\
YOLOV11                             & 0.95  & 640       & 64    & 1e-2  & 5e-4  & 0.1   \\
VGG11 (did not converge)            & 0.64    & 100       & 32    & 1e-2  & 5e-4  & 0.2   \\
ViT\_b16                            &9.2    & 320       & 32    & 3e-4  & 1e-4  & 0.1   \\
\bottomrule
\end{tabular}
\caption{Training cost and hyperparameters across models (BS = batch size; LR = learning rate; WD = weight decay; DR = dropout rate; training time in days).}
\label{tab:model_p}
\end{table*}

\subsection{Geographic Distribution Analysis}

Figure~\ref{fig:contiguous_us} maps the per-tract absolute percentage error, $\left|\hat{y} - y\right|/y$, across the contiguous United States for MambaPop, YOLOv11, and ViT-B/16, with gray tracts denoting ground-truth zero population (excluded from the error calculation). MambaPop (panel~a) produces the best prediction, while YOLOv11's performance (panel~b) is inferior due to the loss of prediction accuracy on large-area tracts (Figure~\ref{fig:large_area_error}).
ViT-B/16 (panel~c) shows markedly higher errors, the map shifting toward saturation nationwide---consistent with its failure to generalize temporally to the 2020 boundaries (Table~\ref{tab:model_m}, 2020 row) and its inability to reach high-population tracts (Figure~\ref{fig:scatter_residual}f).

Figure~\ref{fig:state_error_examples} (Appendix~\ref{app:per_state_maps}) shows per-tract absolute percentage errors for four representative states and geographical regions. Across New Jersey, Massachusetts, Southern California, and Central Washington, the large majority of tracts fall at the low end of the error scale, with high-error tracts concentrated in sparsely populated areas, where small ground-truth populations make the relative error sensitive to modest absolute deviations (for example, in the lightly populated Channel Islands off Southern California).

\begin{figure*}[!htbp]
    \centering
    \includegraphics[width=0.95\textwidth]{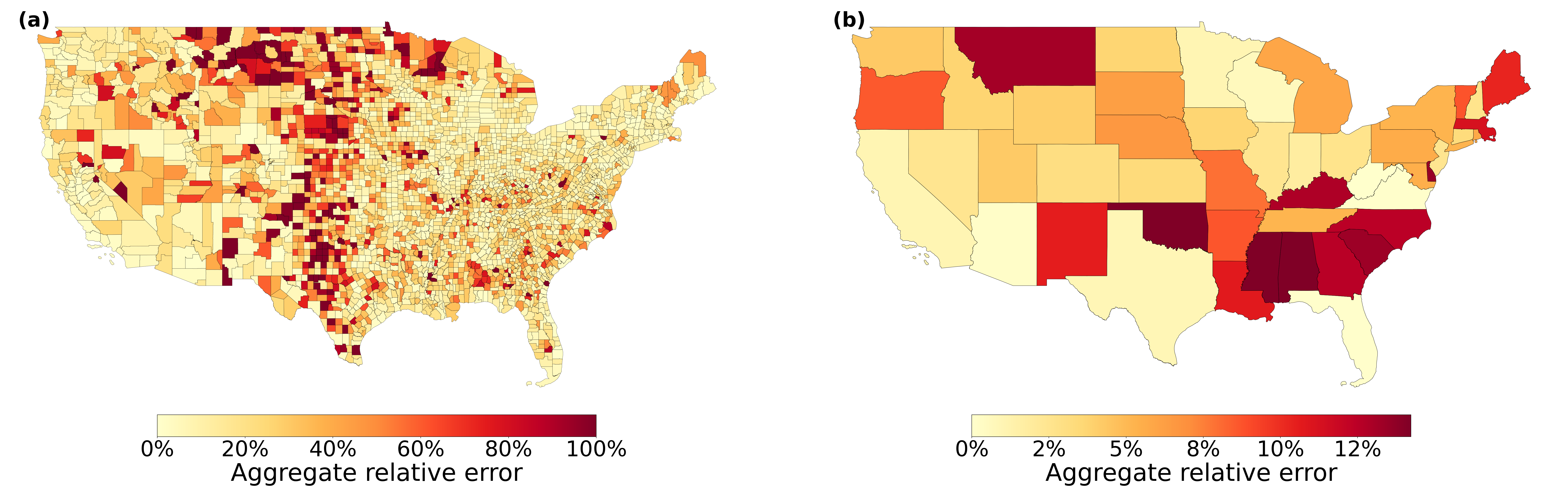}
    \caption{MambaPop aggregate relative error, $\left|\sum\hat{y}-\sum y\right|/\sum y$, across the contiguous United States at (a) county and (b) state levels on the 2020 evaluation set.
    Here, the sums extend over all tracts (with predictions) that fall within a given administrative unit.
    Each panel uses its own color scale (YlOrRd); gray denotes units with no predicted tracts.
    }
    \Description{state_county}
    \label{fig:state_county_error}
\end{figure*}

Figure~\ref{fig:state_county_error} extends this view to the entire contiguous United States, reporting MambaPop's aggregate relative error
at (a) county and (b) state levels. Aggregation sharply reduces the relative error as over- and under-predictions cancel within larger administrative units: the median error is 16\% at the county level, falling to 5.2\% at the state level (and reaching at most 20\% for any state). Thus, MambaPop yields reliable population totals at the regional scales that are most relevant to planning, even if individual tract predictions sometimes result in sizable errors.

\subsection{County-Level Change Validation (2010 \texorpdfstring{$\rightarrow$}{->} 2020)}
\label{sec:county_change}

A central concern for any tract-level estimator trained on a single decennial is whether it has learned features that generalize across census vintages, or has merely memorized 2010 tract identifiers. To test this, we compare predicted and observed changes in the mean tract population between 2010 and 2020 at the \emph{county} level. Counties (state $+$ county Federal Information Processing Standards (FIPS) codes) are stable across decennials, whereas tract GEOIDs are routinely split, merged, or renumbered, making any tract-level temporal extrapolation less reliable. For each model, we aggregate per-tract predictions to per-county means in each year, with
roughly 12{,}000 predictions on the 2010 validation tracts and 84{,}000 predictions on the 2020 tracts.
Then, for each county, we compare its mean tract population in 2010 with that in 2020, keeping only the counties that are present in both years. This procedure results in $n = 1{,}352$ counties in the
comparison dataset.

\begin{figure*}[htbp]
    \centering
    \includegraphics[width=0.95\textwidth]{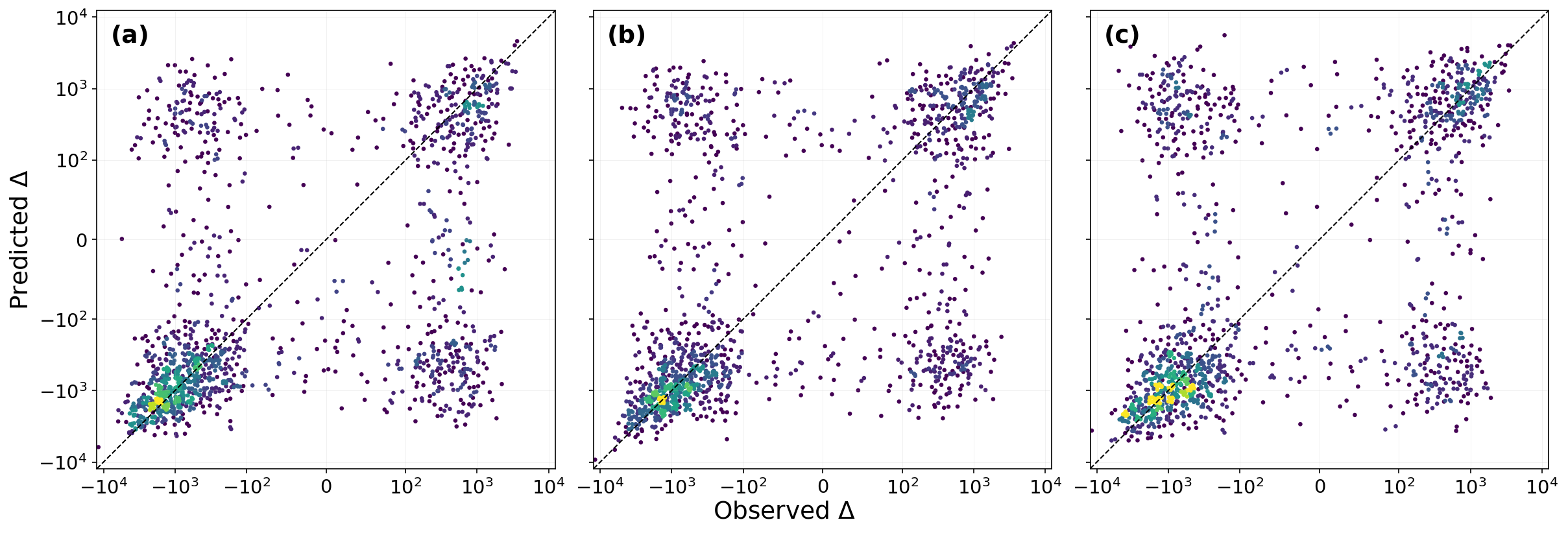}
    \caption{County-level temporal-change validation. Each point is one of the $n = 1{,}352$ counties with predictions in both the 2010 and 2020 datasets; the $x$-axis is the ground-truth change in a county's mean tract population, $\Delta_{\text{actual}} = \overline{\text{pop}}_{2020} - \overline{\text{pop}}_{2010}$, where $\overline{\text{pop}}_{y}$ is the mean population over the county's tracts in year $y$; the $y$-axis is the model-predicted change $\Delta_{\text{pred}}$, on symmetric-log axes ($\text{linthresh} = 100$); the dashed line is $y = x$. Per-model Pearson correlation $r$, ordinary-least-squares (OLS) slope, and mean absolute error (MAE, in persons): (a) MambaPop $r = 0.56$, slope $0.47$, MAE $770$; (b) YOLOv11 $r = 0.62$, slope $0.56$, MAE $752$; (c) ViT-B/16 $r = 0.52$, slope $0.53$, MAE $862$.
    }
    \Description{year difference}
    \label{fig:county_change}
\end{figure*}

As shown in Figure~\ref{fig:county_change}, all three architectures track the direction of inter-decennial change, with the Pearson correlation coefficient $r$ between predicted and observed county-level change of $0.56$ for MambaPop, $0.62$ for YOLOv11, and $0.52$ for ViT-B/16. The models recover the sign of the change for roughly $70\%$ of the counties; the four distinct clusters visible in Figure~\ref{fig:county_change} correspond to four sign combinations of observed and predicted change.
Note that the large majority of counties fall in the two agreement groups, predominantly the shared decrease that reflects the broad decline in mean tract populations from 2010 to 2020. Since the OLS slope relating predicted to observed change is near $0.5$ across all three models ($0.47$, $0.56$, $0.53$), the models capture the direction of change while underestimating its magnitude. This reflects uniform attenuation of the classify-then-convert design rather than a weakness of any single backbone.

The success in recovering temporal population changes reinforces our administrative-unit-native framing: because tract identifiers do not carry across the decennial boundary, a model that has merely memorized 2010 tract geometries could not track the 2020 change.
We conclude that all three models have instead captured population-relevant visual structures. Notably, ViT-B/16 tracks the \emph{relative} change fairly accurately despite failing to generalize in \emph{absolute} terms (its 2020 $R^2$ is negative, Table~\ref{tab:model_m}): it recovers which counties grew or shrank, even though its absolute population estimates remain biased.

\section{Limitations and Future Work}
\label{sec:limitations}

Several limitations remain to be addressed, each pointing to a direction of future work. First, training is capped at the tracts whose images fit within a $512\times512$ crop, excluding the $\sim$14{,}000 largest tracts (19\% of all CONUS census tracts)---predominantly large-area rural and consolidated urban tracts---from the training set. Although MambaPop still tracks population on these out-of-distribution tracts at inference (Figure~\ref{fig:large_area_error}), training at full resolution on the largest tracts, rather than capping the input size for computational efficiency, is the natural next step. Second, we use only the RGB bands (Bands 1, 2, 3) of Landsat 7 to simulate a minimal-input condition; the near-infrared and shortwave-infrared bands---which underlie the Normalized Difference Vegetation Index (NDVI) and built-up-area indices known to correlate with population---are unused, and incorporating them would be a straightforward extension. Third, the present work focuses on the contiguous United States; international validation, particularly in regions where census infrastructure is sparse, is a critical follow-up direction. Finally, extending the framework to continuous temporal monitoring from multi-year satellite imagery would enable tracking population changes over user-specified time periods, without being limited to decennial census cycles.

\section{Conclusion}
\label{sec:conclusion}

In this work, we have introduced MambaPop, a deep-learning architecture that treats population estimation as a sequence-modeling problem: each administrative unit is represented as a polygon-masked image and modeled as a sequence of its own image patches, paired directly with its true SF1 population label. Our approach eliminates the disaggregation and aggregation steps that grid-based methods rely on. Built on the hybrid convolutional--state-space--attention MambaVision backbone, it is, to our knowledge, the first hybrid state-space (Mamba) model applied to the population estimation task.

On all $\sim$84{,}000 contiguous-United-States census tracts of the 2020 census, MambaPop equals the strongest convolutional baseline (YOLOv11) on standard tracts and, alongside it, preserves predictive capability across the inter-decennial gap---where a pure-attention Vision Transformer's absolute estimates collapse---while proving the most robust of the three backbones on the oversized tracts withheld from training, all at a fraction of the transformer's training cost. These results show that sequence modeling supplies the right inductive bias for administrative-unit-native population estimation between decennial census cycles, offering a low-cost complement to in-person enumeration and producing reliable population predictions for the time points that lack authoritative SF1 data.

\begin{acks}
A.V.M. and J.Y. acknowledge financial and logistical support from the Center for Quantitative Biology, Rutgers University.
\end{acks}

\bibliographystyle{ACM-Reference-Format}
\bibliography{references}

\appendix
\newpage
\section{Population Binning Details}
\label{app:binning}

Table~\ref{tab:binning} lists the 18 population classes produced by the $k$-means binning described in Section~\ref{sec:dataset}, together with the number of tracts in each class.

\begin{table}[h]
\centering
\caption{Population classes from one-dimensional $k$-means clustering of 2010 SF1 tract counts ($K=18$). Ranges are in persons; $N_{tracts}$ is the number of contiguous-US tracts in each class.}
\label{tab:binning}
\begin{tabular}{crr}
\toprule
Class & Population range & $N_{tracts}$ \\
\midrule
0  & 0--725            & 960     \\
1  & 726--1{,}625      & 2{,}940 \\
2  & 1{,}626--2{,}167  & 4{,}861 \\
3  & 2{,}168--2{,}649  & 6{,}087 \\
4  & 2{,}650--3{,}109  & 6{,}894 \\
5  & 3{,}110--3{,}567  & 7{,}391 \\
6  & 3{,}568--4{,}038  & 7{,}802 \\
7  & 4{,}039--4{,}534  & 7{,}652 \\
8  & 4{,}535--5{,}074  & 7{,}028 \\
9  & 5{,}075--5{,}682  & 6{,}457 \\
10 & 5{,}683--6{,}374  & 5{,}427 \\
11 & 6{,}375--7{,}200  & 4{,}294 \\
12 & 7{,}201--8{,}317  & 2{,}753 \\
13 & 8{,}318--10{,}096 & 1{,}326 \\
14 & 10{,}101--13{,}049 & 462    \\
15 & 13{,}076--17{,}172 & 157    \\
16 & 17{,}533--23{,}686 & 36     \\
17 & 25{,}039--37{,}452 & 12     \\
\bottomrule
\end{tabular}
\end{table}

\section{Representative Per-Tract Error Maps}
\label{app:per_state_maps}
\begin{figure*}[htbp]
    \centering
    \includegraphics[width=\textwidth]{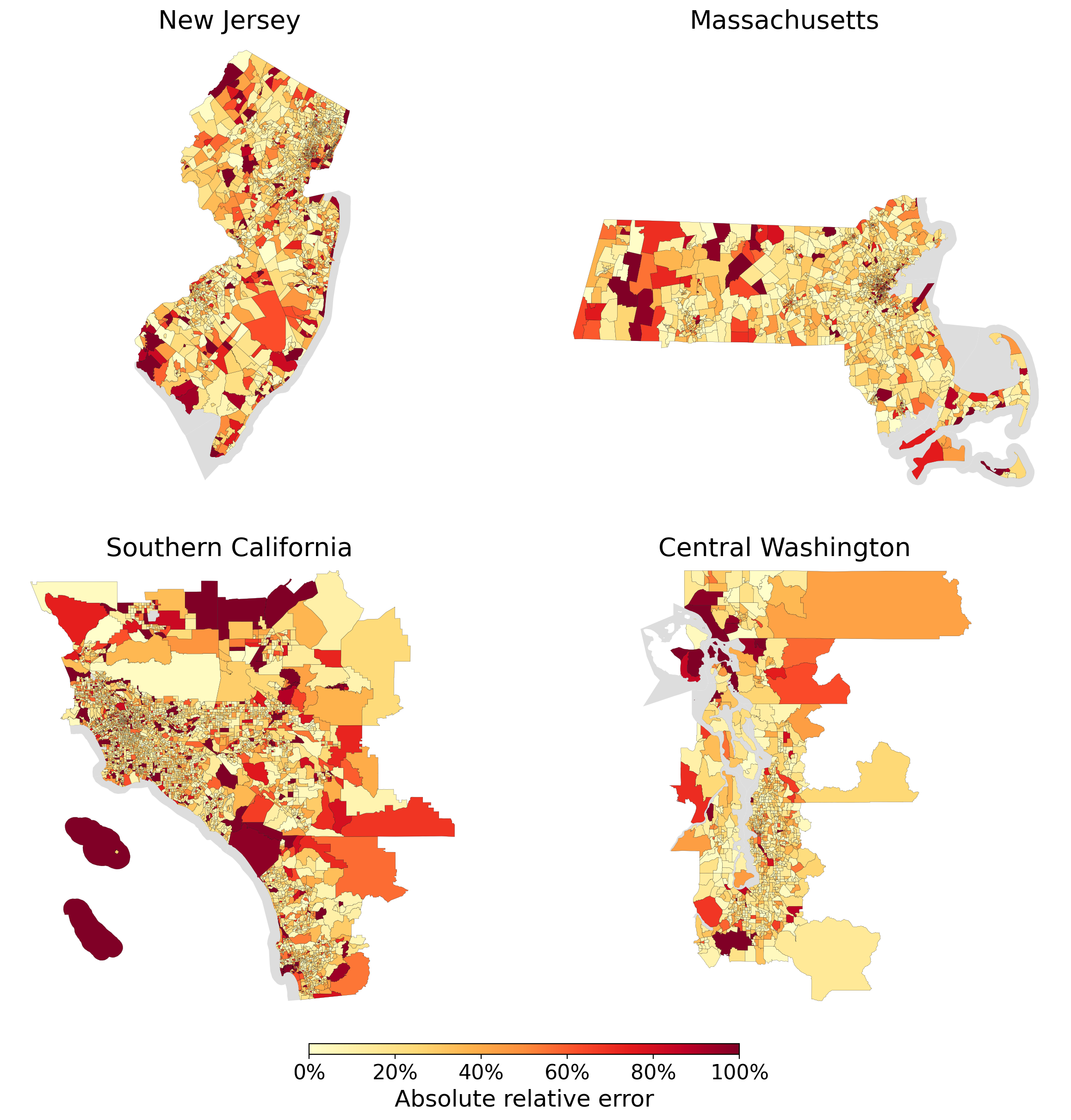}
    \caption{Per-tract absolute percentage error, $\left|\hat{y}-y\right|/y$ (where $y$ and $\hat{y}$ are observed and predicted populations, respectively), of MambaPop predictions for four representative states and geographical regions (New Jersey, Massachusetts, Southern California, Central Washington) on the 2020 census-tract boundaries. Color runs from 0\% to 100\% (YlOrRd); gray denotes tracts with no prediction or zero ground-truth population.}
    \Description{states_error}
    \label{fig:state_error_examples}
\end{figure*}

\end{document}